\pdfoutput=1
\documentclass{article}
\usepackage{arxiv}
\usepackage{times}
\IfFileExists{pcrr7t.tfm}{}{}
\usepackage{natbib}
\setcitestyle{authoryear,round,citesep={;},aysep={,},yysep={;}}
\usepackage{amsmath,amssymb,amsthm}
\usepackage{enumitem}
\usepackage{algorithm,algorithmic}
\usepackage{booktabs}
\usepackage{multirow}
\usepackage[table]{xcolor}
\usepackage{graphicx}
\graphicspath{{figures/}}
\usepackage{tcolorbox}
\tcbuselibrary{listings,breakable}
\newtcblisting{promptbox}[1]{listing only, breakable, colback=black!3, colframe=black!40, boxrule=0.4pt, arc=1.5pt, left=4pt, right=4pt, top=2pt, bottom=2pt, title={#1}, fonttitle=\bfseries\footnotesize, listing options={basicstyle=\ttfamily\footnotesize, breaklines=true, columns=fullflexible, keepspaces=true, breakindent=0pt, upquote=true, literate={"}{{\char34}}1}}
\usepackage{amsmath,amsfonts,bm}

\def\eqref#1{equation~\ref{#1}}

\def\1{\bm{1}}

\DeclareMathAlphabet{\mathsfit}{\encodingdefault}{\sfdefault}{m}{sl}
\SetMathAlphabet{\mathsfit}{bold}{\encodingdefault}{\sfdefault}{bx}{n}

\definecolor{ourrow}{RGB}{226,230,247}
\newcommand{\ours}{\rowcolor{ourrow}}
\newcommand{\na}{--}
\newcommand{\sub}[1]{\hspace{1em}#1}

\newif\ifshowalgorithm
\showalgorithmtrue
\newif\ifshowappendix
\showappendixtrue

\theoremstyle{remark}

\usepackage{hyperref}
\usepackage{url}
\hypersetup{pdftitle={Learning to Steer LLM Agents and Multi-Agent Systems}}

\title{Learning to Steer LLM Agents and Multi-Agent Systems}

\newcommand{\authorgap}{\ \ \ \ \ }
\author{%
  Leyang Shen$^{1,3}$\authorgap Yaxin Gong$^{2,3}$\authorgap Yang Zhang$^{1*}$\authorgap Chenxiao Fan$^{2,3}$\authorgap Jiakai Wang$^{3}$ \\
  \bfseries Dong Wang$^{3}$\authorgap Yang Liu$^{3}$\authorgap Fuli Feng$^{2}$\authorgap Chun Kai Ling$^{1}$\authorgap Tat-Seng Chua$^{1}$ \\[3pt]
  $^{1}$ National University of Singapore \\
  $^{2}$ University of Science and Technology of China \quad $^{3}$ Qwen Business Unit of Alibaba \\[3pt]
  \small $^{*}$ Corresponding author \\
  \small\texttt{lshen@u.nus.edu}
}

\makeatletter
\DeclareRobustCommand{\novspace}{\@ifstar\@gobble\@gobble}
\newsavebox{\shrinkbox}
\let\origresizebox\resizebox
\newcommand{\shrinkresizebox}[3]{\leavevmode\sbox{\shrinkbox}{#3}\ifdim\wd\shrinkbox>#1\relax\origresizebox{#1}{#2}{\usebox{\shrinkbox}}\else\usebox{\shrinkbox}\fi}
\newcommand{\inputarxiv}[1]{{\let\vspace\novspace\let\resizebox\shrinkresizebox\input{#1}}}
\makeatother

\begin{document}

\maketitle

\begin{abstract}

Large language model (LLM)-based agents and multi-agent systems increasingly tackle complex tasks through long-horizon tool use and inter-agent collaboration. However, these systems are prone to deviating from the overall goal as execution unfolds, which wastes execution resources and leads to undesired outcomes, a failure termed \emph{goal drift}. In this work, we propose a steering agent that supervises execution on the fly and steers the system back on track when drift occurs. The steering agent observes the execution context, deciding when to intervene with a low-cost fast path and how to intervene with a high-cost slow path. To unlock its steering intelligence, the ability to assess goal drift and generate effective interventions, we jointly train the two paths with reinforcement learning, combining a pairwise ranking objective with group-relative policy optimization (GRPO). Experiments on information-seeking search tasks show that the steering agent provides effective supervision for both single-agent and multi-agent systems, improving their performance and efficiency. The learned steering intelligence generalizes to out-of-domain tasks, transfers to unseen executors, and supports controllable intervention frequency without retraining.

\end{abstract}

\section{Introduction}
\vspace{-0.3em}

Large language model (LLM)-based agents~\citep{yao2023react,wang2024agentsurvey,xi2025agentsurvey} and multi-agent systems (MAS)~\citep{wu2023autogen,guo2024multiagents,li2024multiagents} increasingly tackle complex tasks through long-horizon tool use~\citep{schick2023toolformer} and inter-agent collaboration~\citep{hong2024metagpt}, spanning information seeking~\citep{tongyideepresearch}, software engineering~\citep{codex}, and scientific discovery~\citep{coral,edgebench}. However, these systems are prone to \emph{goal drift}, where agents deviate from the overall goal as execution unfolds~\citep{arike2025technical,cemri2026multi,laban2026llms}. A single agent may persist along an unproductive chain of actions, while an agent in a MAS may prioritize its local objective over the collective goal. \emph{Goal drift} can waste execution resources and lead to undesired outcomes, even when individual actions appear locally reasonable~\citep{cemri2026multi,liu2026diagnosing}.

\begin{figure}[b]
    \centering
    \includegraphics[width=0.95\linewidth]{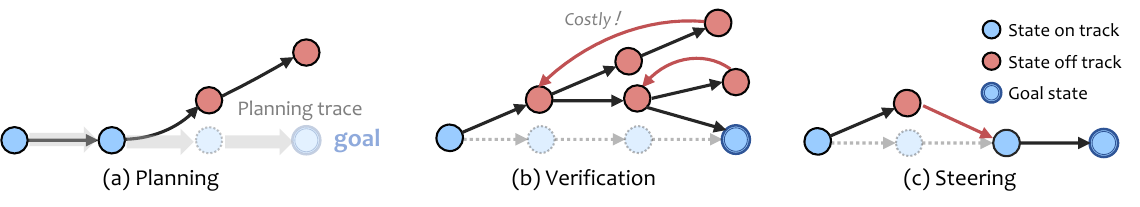}
    \vspace{-1.2em}
    \caption{\textbf{Scheme comparison of planning, verification, and steering.} (a) Execution can deviate from an initial plan. (b) Verification can cause costly retries. (c) Steering provides on-the-fly intervention to keep the system on track.}
    \label{fig:planning_verification_steering}
\end{figure}

To keep agents on track and ensure goals are achieved during long-horizon execution, prevalent agent systems, such as Claude Code~\citep{anthropic2026claudecode} and Codex~\citep{codex}, employ planning and verification~\citep{young2025harnesses,anthropic2026claudecodehooks}. As illustrated in Fig.~\ref{fig:planning_verification_steering}, planning decomposes a goal into manageable subgoals~\citep{wang2023planandsolve}, but execution may still deviate from the planned route. Verification checks whether the goal has been achieved before termination, but such post-hoc remediation can lead to irreversible consequences and costly retries. These limitations motivate a natural design of \emph{on-the-fly supervision}, where an external agent monitors execution and steers the system back toward its goal when drift occurs, complementing planning and verification.

Such supervision is conceptually straightforward but challenging for the LLM backbone. It requires \emph{steering intelligence}, the ability to assess execution against the overall goal, determine when intervention is needed, and generate actions that steer systems back on track. Such a capability is established in other goal-directed systems, such as direct supervision in organizations~\citep{mintzberg1979structuring}, supervisory control of automated systems~\citep{sheridan1992telerobotics}, and metacognitive monitoring and control in human cognition~\citep{nelson1990metamemory}. We bring this principle to LLM agent systems by building a steering agent and improving its steering intelligence through post-training.

Specifically, we define an observation space that exposes the task goal and execution context of the agent system to the steering agent. During execution, each observation update triggers a decision about whether and, if needed, how to intervene. These two demands motivate a natural design with fast and slow paths~\citep{kahneman2011thinking,su2025dualformer,lampson1983hints,clark1989analysis}. The fast path estimates steering priority through a low-cost single-token prediction. A threshold on this priority controls activation of the slow path, which leverages more costly reasoning to generate steering actions.
By separating fast judgment from deliberative generation, this design balances monitoring efficiency and steering effectiveness.
We then improve the steering intelligence of the steering agent by jointly training the two paths with reinforcement learning, combining a pairwise ranking objective~\citep{burges2005ranknet} for fast-path steering judgment with group-relative policy optimization (GRPO)~\citep{shao2024deepseekmath} for slow-path action generation.

We evaluate our steering approach on information-seeking search tasks~\citep{mialon2023gaia, ho2020constructing} in both single-agent and multi-agent settings. Our single-agent experiments show that steering is challenging even for a capable LLM backbone, and that reinforcement learning substantially improves steering intelligence. Across seven question-answering benchmarks, zero-shot steering yields a performance gain of only 1.84 points, whereas the RL-optimized steering agent improves performance by 6.66 points with only 0.93 steering actions per task. Furthermore, the steering intelligence generalizes to out-of-domain tasks, transfers to unseen executors, and supports controllable intervention frequency without retraining, with steering at the highest frequency raising the gain to 17.11 points. In the multi-agent setting, we introduce the steering agent into a five-role multi-agent system~\citep{xu2026tacomas}. In this more challenging steering task, where zero-shot supervision degrades performance, the RL-trained steering agent achieves a gain of 5.34 points.

Our contributions are summarized as follows.
\begin{itemize}[leftmargin=*]
    \item We characterize goal drift in single- and multi-agent systems, where local objectives can displace the overall task goal, and motivate on-the-fly supervision to keep execution on track.
    \item We develop a steering agent with fast intervention judgment and deliberative action generation, and improve its steering intelligence through reinforcement learning.
    \item We conduct experiments on both single- and multi-agent settings, where the steering agent demonstrates effective performance gains, strong generalizability, transferability, and controllability.
\end{itemize}

\section{Related Work}

\subsection{LLM-based Agent Systems}

LLM-based single- and multi-agent systems are increasingly applied to complex, long-horizon tasks, including information-seeking research~\citep{wideresearch, tongyideepresearch}, software engineering~\citep{codex, deepseekharness}, and scientific discovery~\citep{edgebench, coral}.
However, as increasingly complex tasks require longer execution horizons, these systems may deviate from their original task objectives, a failure mode known as \emph{goal drift}~\citep{arike2025technical, cemri2026multi, laban2026llms}.

To better equip LLMs for long-horizon interaction and multi-agent collaboration, recent work has explored agentic post-training, particularly reinforcement learning (RL)~\citep{chen2025loop, wang2025ragen, marft}. Long-horizon agentic RL optimizes LLMs end-to-end with tools on complex tasks requiring extended interaction~\citep{gao2025asearcher, tongyideepresearch, harness1}.
Multi-agent RL jointly optimizes collaborating agents under shared task-level objectives, encouraging individual agents to contribute to overall task success~\citep{marft, stronger-mas}.
However, long-horizon rollouts can make training computationally expensive~\citep{gao2025asearcher}, and RL fine-tuning can overfit to the training environment, affecting the generalization ability of the foundation model~\citep{aissi2024promptoverfitting, liu2026generalizationtax, lv2026openagent}.
In contrast, we keep the task-executing LLMs frozen and train an external agent to supervise execution and steer the system back toward its overall goal when drift occurs.

\subsection{Goal Drift Mitigation}

To mitigate goal drift during long-horizon execution, prevalent agent systems~\citep{deepseekharness,young2025harnesses, codex} employ planning and verification to support reliable execution. Planning decomposes a goal into manageable subgoals that guide subsequent actions~\citep{wang2023planandsolve}. Verification checks whether goals have been met before an agent declares completion, prompting further execution or revision when needed~\citep{anthropic2026claudecodehooks,lopopolo2026harness}.
However, a plan does not ensure reliable execution~\citep{sinha2026illusion}, while verification may uncover deviations after substantial effort has already been spent.

These limitations motivate a natural on-the-fly supervision design, where an external supervisor monitors intermediate execution and intervenes when goal drift occurs.
Recent work explores this idea with a primary focus on detection, identifying when and where an agent goes wrong~\citep{zhang2025which,zhang2026agentforesight,baidya2026evidence}.
SupervisorAgent~\citep{lin2026stop} takes a step toward correction by prompting an external LLM, but it targets easily fixable issues such as errors and repetitive actions. Goal drift, however, is more challenging to detect and correct in real time, as it emerges gradually from locally plausible actions. In this paper, we address this challenge with a steering agent that jointly learns drift detection and correction through reinforcement learning.

\section{Learning to Steer LLM-based Agent Systems}
\label{sec:method}

In this section, we begin by discussing goal drift that happens in both single-agent and multi-agent systems (Section~\ref{sec:goal_drift}). Then, we introduce our steering agent design (Section~\ref{sec:protocol}) and the reinforcement learning algorithm (Section~\ref{sec:learning}).

\subsection{Goal Drift in LLM-based Agent Systems}
\label{sec:goal_drift}

An LLM-based agent executes a task with an overall goal $g$ by interacting with an environment~\citep{yao2023react}. At each step, the agent generates an action conditioned on $g$ and the interaction history, and receives environment feedback. A multi-agent system distributes this process across specialized agents, where each agent handles a temporary subgoal derived from $g$, such as retrieving evidence or verifying an answer, and hands its output to subsequent agents. In both settings, execution unfolds as a sequence of temporary subgoals in service of the overall goal~\citep{march1958organizations}.

In LLM-based agent systems, \emph{goal drift} appears when a temporary subgoal dominates the overall task goal~\citep{arike2025technical}, leading to unproductive actions that might be locally reasonable. Goal drift reduces execution efficiency by consuming resources on actions that contribute little to task completion. If the deviation remains uncorrected, the agent system may never finish the task or finish it with an undesired outcome.

\begin{figure}[t]
    \centering
    \vspace{-1.5em}
    \includegraphics[width=\linewidth]{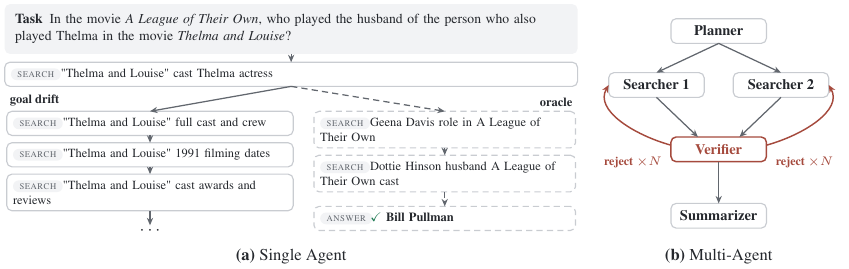}
    \vspace{-1.5em}
    \caption{\textbf{Goal drift example in single-agent and multi-agent executions.} (a) The agent keeps investigating the same topic after obtaining enough information, wasting execution on unproductive actions. (b) The verifier repeatedly rejects answers the searchers have already delivered, sending execution into repeated revision rounds.}
    \label{fig:goal_drift_cases}
    \vspace{-1em}
\end{figure}

For example, in multi-hop search tasks~\citep{ho2020constructing,yang2018hotpotqa}, the agent needs to retrieve several distinct pieces of evidence and combine them to answer a question, which involves a sequence of temporary retrieval subgoals. A common failure mode is that the agent keeps searching over the same topic when key information has already been obtained.
As shown in Fig.~\ref{fig:goal_drift_cases}(a), the first search already identifies the actress required to answer the question.
The agent spends three more searches on the same film instead of moving on to the next step, consistent with observed patterns of redundant and unproductive search~\citep{liu2026diagnosing}.

In a multi-agent system, where specialized agents collaborate on a complex task, goal drift arises similarly when an agent prioritizes its role objective over the overall task goal. A commonly observed failure mode is over-rejection, where a verifier falsely rejects results that already satisfy the task goal~\citep{huang2024selfcorrect,stechly2025selfverification,xu2025tinyv}. As shown in Fig.~\ref{fig:goal_drift_cases}(b), the verifier repeatedly rejects the answers the searchers have delivered, leading execution to repeated revision rounds that can turn a successful execution into a failure. This reflects the verifier overweighting its own goal of admitting only answers beyond doubt.

\subsection{The Steering Agent}
\label{sec:protocol}

To mitigate goal drift, we design a \emph{steering agent} that supervises agent execution on the fly and steers it back on track before further effort is wasted, achieving efficient goal completion.

\paragraph{Steering Interface.}

Consider an agent system executing a task with goal $g$. We define an observation space $\mathcal{O}$ through a function $O$ that exposes the goal and available execution context, and an action space $\mathcal{A}_{\mathrm{steer}}=\{\mathrm{pass}\}\cup\mathcal{A}_{\mathrm{int}}$, where $\mathrm{pass}$ lets execution continue and $\mathcal{A}_{\mathrm{int}}$ contains the interventions supported by the system. At steering decision $t$, the steering agent reads the observation $o_t=O(g,s_t)\in\mathcal{O}$, where $s_t$ is the underlying execution state that collects the goal and the interaction so far, and returns a steering action:
\begin{equation}
    a_t\sim\pi_\theta\left(\cdot\mid o_t\right),
    \label{eq:steering_task}
\end{equation}
where the steering policy $\pi_\theta$, realized by an LLM with parameters $\theta$, maps each observation to a distribution over $\mathcal{A}_{\mathrm{steer}}$. The system applies the sampled action $a_t$ before resuming execution.
This steering interface can achieve different steering granularities depending on the observation and action spaces. For example, exposing every execution step of a single agent enables fine-grained supervision, while exposing only inter-agent handoffs enables coarse-grained supervision over a multi-agent system.

We formalize the steering process as a partially observable Markov decision process (POMDP)~\citep{sutton2018reinforcement,kaelbling1998planning}, where the agent system being supervised acts as the environment. At decision step $t$, the steering agent observes $o_t$ rather than the full execution state $s_t$, takes the action $a_t$, and the agent system transitions from $s_t$ to
\begin{equation}
    s_{t+1}\sim\mathcal{T}\!\left(\cdot\mid s_t,a_t\right),
    \label{eq:steering_mdp}
\end{equation}
where the transition $\mathcal{T}$ is realized by the system resuming execution until the next observation update. The task reward $R$ is received when execution terminates and measures goal completion.

\begin{figure}[t]
    \centering
    \vspace{-2em}
    \includegraphics[width=0.95\linewidth]{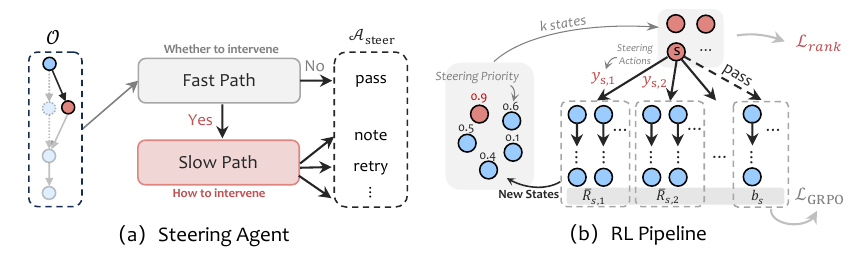}
    \vspace{-1em}
    \caption{\textbf{Steering agent and reinforcement learning pipeline.} (a) The steering agent decides whether to intervene with the fast path and how to intervene with the slow path. (b) The RL pipeline samples states from previous rollouts, samples several steering actions at each selected state, and jointly updates the two paths with a pairwise ranking objective and GRPO.}
    \label{fig:protocol}
    \vspace{-0.8em}
\end{figure}

As shown in Figure~\ref{fig:protocol}(a), each observation update marks a state transition in observation space and triggers a steering decision. The steering agent then judges whether the current state needs intervention and generates the intervention when it does. These two demands split the agent into a fast path and a slow path~\citep{kahneman2011thinking,su2025dualformer,lampson1983hints,clark1989analysis}. The fast path judges the need for intervention at low cost, and the slow path produces the reasoning and action for intervention. Both paths share the same LLM backbone parameterized by $\theta$.

\paragraph{Fast path.} To process each observation update at low cost, the fast path judges the need for intervention within a single LLM forward pass. It operates by estimating the \emph{steering priority} of the current state, which measures how much a steering action at this state can change the final outcome. This allows the steering agent to achieve a larger impact with a fixed intervention budget.
Specifically, the fast path measures steering priority $v_\theta(o_t)$ by a \texttt{yes}/\texttt{no} judgment, which is calculated by reading the probabilities of the two tokens:
\begin{equation}
    z_\theta(o_t)=\log p_\theta(\texttt{yes}\mid P_{\mathrm{fast}}(o_t))-\log p_\theta(\texttt{no}\mid P_{\mathrm{fast}}(o_t)),
    \qquad v_\theta(o_t)=\sigma(z_\theta(o_t)),
    \label{eq:judge_score}
\end{equation}
where $P_{\mathrm{fast}}(o_t)$ denotes the fast-path LLM input built from $o_t$, $p_\theta$ is the token probability given by the shared LLM backbone, $z_\theta(o_t)$ is the log-odds between the two tokens, and the logistic function $\sigma$ normalizes it to steering priority $v_\theta(o_t)\in(0,1)$.

We then follow \citet{bengio2013estimating} to trigger the slow path with a soft threshold on the priority:
\begin{equation}
    I_t\sim\operatorname{Bernoulli}(p_t),
    \qquad
    p_t=\sigma\!\left(\operatorname{logit}(v_\theta(o_t))-\operatorname{logit}(\tau)\right),
    \label{eq:steering_gate}
\end{equation}
where $I_t\in\{0,1\}$ indicates whether the slow path is triggered and $\tau\in(0,1)$ is the trigger threshold. The function $\operatorname{logit}(x)=\log\frac{x}{1-x}$ is the inverse of the logistic function $\sigma$, mapping the priority and the threshold onto the log-odds scale. The soft threshold is a smoothed version of the hard rule $\mathbf{1}[v_\theta(o_t)>\tau]$, whose trigger probability $p_t$ exceeds one half once the priority exceeds $\tau$ and approaches one as the gap grows.
We write $\pi_\theta^{\mathrm{fast}}(I_t\mid o_t)$ for this trigger distribution, which represents the probability of entering the slow path given the observation $o_t$.

\paragraph{Slow path.} Once triggered, the slow path conducts chain-of-thought reasoning~\citep{wei2022chain} and generates a steering action. Let $r_t$ denote the generated reasoning and $a_t\in\mathcal{A}_{\mathrm{int}}$ the resulting steering action, produced as one output sequence $y_t=(r_t,a_t)$. The slow path can then be represented by an LLM auto-regressive generation process:
\begin{equation}
    \begin{aligned}
        \pi_\theta^{\mathrm{slow}}(r_t,a_t\mid o_t)
        &=p_\theta(r_t\mid P_{\mathrm{slow}}(o_t))\,p_\theta(a_t\mid P_{\mathrm{slow}}(o_t),r_t)\\
        &=\prod_{\ell=1}^{|y_t|}p_\theta(y_{t,\ell}\mid P_{\mathrm{slow}}(o_t),y_{t,<\ell}),
    \end{aligned}
    \label{eq:slow_policy}
\end{equation}
where $P_{\mathrm{slow}}(o_t)$ denotes the slow-path input built from $o_t$. The LLM first generates the reasoning $r_t$ and then commits to the action $a_t$ conditioned on that reasoning. The second line expands this process at the token level, where each token $y_{t,\ell}$ is generated auto-regressively conditioned on the preceding tokens $y_{t,<\ell}$. Through the transition in Equation~\ref{eq:steering_mdp}, the steering action changes the execution state and thereby the actions the executor takes next, steering the system back on track. Execution of the agent system being supervised then resumes until the next observation update.

We write $\pi_\theta^{\mathrm{slow}}(a_t\mid o_t)=\sum_{r}\pi_\theta^{\mathrm{slow}}(r,a_t\mid o_t)$ for the action distribution of the slow path, marginalizing over the reasoning. Combining the two paths then gives the steering policy in Equation~\ref{eq:steering_task}:
\begin{equation}
    \pi_\theta(a_t\mid o_t)=\pi_\theta^{\mathrm{fast}}(I_t{=}0\mid o_t)\,\mathbf{1}[a_t=\mathrm{pass}]+\pi_\theta^{\mathrm{fast}}(I_t{=}1\mid o_t)\,\pi_\theta^{\mathrm{slow}}(a_t\mid o_t),
    \label{eq:steering_policy}
\end{equation}

where the fast path passes with probability $1-p_t$ and otherwise delegates to the slow path.

\subsection{Reinforcement Learning for Steering Intelligence}
\label{sec:learning}

Effective steering requires steering intelligence, the ability to assess execution against the overall goal, determine when intervention is needed, and generate actions that bring execution back on track. To unlock this capability for the LLM backbone, we conduct post-training on the steering interface, jointly optimizing the two paths with feedback from task outcomes.

As illustrated in Figure~\ref{fig:protocol}(b), the reinforcement learning pipeline samples intermediate states that previous rollouts have reached and lets the steering agent attempt interventions there. The steering agent generates several steering actions at the sampled states, and their final rewards are compared together with those of the unsteered continuations. The comparison teaches the fast path which states carry high steering priority through a pairwise ranking objective, and teaches the slow path which actions improve outcomes through GRPO~\citep{shao2024deepseekmath}. Training keeps the executor frozen and updates only the parameters $\theta$ of the steering agent.

\paragraph{Training pipeline.} For each task, an unsteered rollout of the frozen executor initializes a state bank containing the states reached during execution, and the fast path computes a steering priority $v_\theta(o_s)$ for every state $s$ added to the bank with Equation~\ref{eq:judge_score}, where $o_s=O(g,s)$ denotes the observation of state $s$. The pipeline then performs $m$ rounds of state selection without replacement. In each round, we select up to $k$ states, where each selection takes the state with the highest steering priority with probability $1-\varepsilon_{\mathrm{sel}}$ and a uniformly random state otherwise, balancing exploitation and exploration. At each selected state $s$, we sample $G$ outputs $y_{s,j}=(r_{s,j},a_{s,j})$ from the slow policy in Equation~\ref{eq:slow_policy}.
We apply each steering action to the selected state $s$, yielding the steered states $\{s'_j\}_{j=1}^{G}$. To reduce variance, the executor continues the rollout $N$ times from every state in $\{s\}\cup\{s'_j\}_{j=1}^{G}$, and the mean reward over the continuations of each steering action serves as the reward of that action:
\begin{equation}
    \bar R_{s,j}=\frac{1}{N}\sum_{n=1}^{N}R_{s,j,n},
    \qquad
    b_s=\frac{1}{N}\sum_{n=1}^{N}R^{0}_{s,n},
    \label{eq:outcome_stats}
\end{equation}
where $R_{s,j,n}$ is the task reward of the $n$-th continuation from the steered state $s'_j$, and $R^{0}_{s,n}$ that of the $n$-th continuation from the unsteered state $s$, averaged into the baseline $b_s$.
This rollout pipeline provides controlled comparisons among steering actions and the unsteered baseline, allowing the steering agent to learn which states merit intervention and which actions improve outcomes.

\paragraph{Rank-based fast-path learning.} The fast path learns to rank states by steering priority. As the training target, we measure the priority of state $s$ by the mean absolute outcome change that steering causes relative to the no-intervention baseline,
\begin{equation}
    q(s)=\frac{1}{G}\sum_{j=1}^{G}\left|\bar R_{s,j}-b_s\right|.
    \label{eq:sensitivity_target}
\end{equation}
That is, $q(s)$ averages how far the $G$ steered outcomes $\bar R_{s,j}$ move from the unsteered baseline $b_s$. For the evaluated states $\mathcal{S}=\{s_i\}$ from the same task, we write $z_i=z_\theta(o_{s_i})$ for the fast-path score of state $s_i$ in Equation~\ref{eq:judge_score} and $q_i=q(s_i)$ for its measured priority. The set $\mathcal{P}=\{(i,j)\mid i<j,\ q_i\ne q_j\}$ collects state pairs with different priorities, and $d_{ij}=\mathbf{1}[q_i>q_j]$ indicates which state ranks higher.
We apply the pairwise ranking loss~\citep{burges2005ranknet} for fast-path learning
\begin{equation}
    \mathcal{L}_{\mathrm{rank}}=-\frac{1}{\max(|\mathcal{S}|-1,1)}
    \sum_{(i,j)\in\mathcal{P}}\left[d_{ij}\log\sigma(z_i-z_j)+(1-d_{ij})\log\sigma(z_j-z_i)\right].
    \label{eq:judge_rank}
\end{equation}
The loss pushes $z_i$ above $z_j$ whenever state $i$ shows a higher measured priority, aligning the predicted ordering with the measured one, and tied pairs contribute no signal.

\paragraph{GRPO-based slow-path learning.} We optimize the slow path with GRPO~\citep{shao2024deepseekmath} over the multiple steering attempts sampled at the same state. To push steering actions toward positive improvement, we include the unsteered baseline $b_s$ in the within-group normalization. The advantage of each steering action is calculated by:
\begin{equation}
    U_s=\{\bar R_{s,1},\ldots,\bar R_{s,G},b_s\},
    \qquad
    A^{\mathrm{act}}_{s,j}=\frac{\bar R_{s,j}-\operatorname{mean}(U_s)}{\operatorname{std}(U_s)}.
    \label{eq:note_advantage}
\end{equation}
That is, an action receives a positive advantage only when its mean reward exceeds the pooled average of the steered outcomes and the unsteered baseline in $U_s$. We assign the advantage to every generated reasoning and action token and apply the standard PPO loss~\citep{schulman2017ppo}:
\begin{equation}
    \mathcal{L}_{\mathrm{GRPO}}=
    -\frac{1}{|\mathcal{D}|}
    \sum_{(o,y,A)\in\mathcal{D}}
    \frac{1}{|y|}\sum_{\ell=1}^{|y|}
    \min\Big[
        \rho_\ell(\theta)A,\
        \operatorname{clip}\big(\rho_\ell(\theta),1-\epsilon,1+\epsilon\big)A
    \Big],
    \label{eq:grpo_loss}
\end{equation}
where $\mathcal{D}$ collects the sampled outputs $(o,y,A)$ across states with their advantages, $\rho_\ell(\theta)=p_\theta(y_\ell\mid P_{\mathrm{slow}}(o),y_{<\ell})/p_{\theta_{\mathrm{old}}}(y_\ell\mid P_{\mathrm{slow}}(o),y_{<\ell})$ is the token probability ratio against the sampling policy $\pi_{\theta_{\mathrm{old}}}$, and $\epsilon$ is the clipping range.

Combining the objectives of the fast and slow paths, we obtain the training loss
\begin{equation}
    \mathcal{L}=\mathcal{L}_{\mathrm{GRPO}}+\mathcal{L}_{\mathrm{rank}},
    \label{eq:joint_loss}
\end{equation}
which jointly updates the shared parameters $\theta$.

\section{Experiments}
\label{sec:experiments}

In this section, we conduct experiments to study three research questions.
\textbf{RQ1:} Can a steering agent improve task performance, and can reinforcement learning strengthen its steering intelligence?
\textbf{RQ2:} Does the learned steering capability generalize to out-of-domain tasks, transfer to unseen executors, and support controllable intervention frequency?
\textbf{RQ3:} Can steering improve a multi-agent system, where goal drift arises from collaboration among specialized roles?

\subsection{Experimental Setup}
We follow the experimental setting of ASearcher~\citep{gao2025asearcher}, which trains on ASearcher-Base-35k and evaluates on seven retrieval-augmented QA benchmarks of multi-hop reasoning and single-hop retrieval with a local Wikipedia retrieval server. We report F1~\citep{rajpurkar2016squad} following the same protocol, with Avg. denoting the mean over the seven benchmarks.
For the executor, we use a Qwen2.5-7B-Instruct search agent~\citep{qwen2024qwen25}, which shows basic search ability but is prone to goal drift. Given the difficulty of the steering task, we use the stronger Qwen3-8B~\citep{yang2025qwen3} as the backbone of the steering agent.
Further details are described in Appendix~\ref{app:implementation}.

\subsection{Effectiveness of Steering (RQ1)}

\begin{table}[t]
\centering
\footnotesize
\setlength{\tabcolsep}{3pt}
\vspace{-1.5em}
\caption{\textbf{Results on in-domain benchmarks.} $\Delta$ is the gain in Avg. over No Intervention, and Cost reports steering actions (\#Act.) and executor turns (Turns) per task. Best results are in bold.}
\label{tab:id}
\begin{tabular}{l|cccc|ccc|cc|cc}
\toprule
\multirow{2}{*}{Method} & \multicolumn{4}{c|}{Multi-Hop QA}
 & \multicolumn{3}{c|}{Single-Hop QA}
 & \multicolumn{2}{c|}{Overall}
 & \multicolumn{2}{c}{Cost} \\
\cmidrule(lr){2-5}\cmidrule(lr){6-8}\cmidrule(lr){9-10}\cmidrule(lr){11-12}
 & 2Wiki & Hotpot & Bamboogle & MuSiQue & NQ & Trivia & Pop
 & Avg. & $\Delta$ & \#Act. & Turns \\
\midrule
No Intervention & 36.9 & 46.9 & 38.4 & 20.8 & 46.8 & 60.6 & 48.5 & 42.70 & \na & \na & 6.4 \\
\midrule
Planning & 38.1 & 45.0 & 43.6 & 21.9 & 47.9 & 61.3 & 48.4 & 43.74 & +1.04 & 1.00 & 8.0 \\
\sub{w/ RL} & 46.5 & 49.7 & 45.9 & 22.9 & 47.5 & 61.4 & 47.5 & 45.91 & +3.21 & 1.00 & 8.8 \\
\midrule
Verification & 35.2 & 43.8 & 44.6 & 19.3 & 44.1 & 60.8 & 43.2 & 41.57 & $-1.13$ & 1.22 & 7.8 \\
\sub{w/ RL} & 40.8 & 51.1 & 39.9 & 20.9 & 51.1 & 61.4 & 48.9 & 44.87 & +2.17 & 1.08 & 6.8 \\
\midrule
Steering & 41.9 & 48.2 & 41.4 & 24.0 & 48.9 & 60.5 & 46.9 & 44.54 & +1.84 & 1.11 & 6.5 \\
\sub{w/ Filtered BC} & 43.6 & 46.8 & 38.2 & 22.5 & 47.3 & 61.2 & 48.9 & 44.07 & +1.37 & 0.95 & \textbf{6.2} \\
\ours \sub{w/ RL} & \textbf{50.1} & \textbf{53.7} & \textbf{47.9} & \textbf{26.6} & \textbf{51.9} & \textbf{63.2} & \textbf{52.1} & \textbf{49.36} & \textbf{+6.66} & \textbf{0.93} & \textbf{6.2} \\
\bottomrule
\end{tabular}
\end{table}

To study the effectiveness of the steering agent, we compare it with planning and verification, which use the same supervisor backbone (Qwen3-8B) to decompose the goal before execution and to review the goal completion afterward.
 Table~\ref{tab:id} reports F1 on the seven in-domain benchmarks, together with the corresponding steering cost and execution cost, i.e., steering actions (\#Act.) and executor turns (Turns) per task. We adjust the fast path threshold (Equation~\ref{eq:steering_gate}) to keep the average number of steering actions around one per task for a fair comparison.

Zero-shot steering brings the largest gain among the three supervision forms and leaves executor turns nearly unchanged. In comparison, zero-shot planning and verification help less or even degrade performance and lengthen execution. These results show that intervening during execution supervises the executor more effectively than planning ahead or reviewing afterward.

To examine the improvement brought by reinforcement learning, we apply the same learning algorithm (Section~\ref{sec:learning}) to the steering, planning, and verification agents.
RL improves all three supervision forms over their zero-shot counterparts.
Among them, steering benefits the most. RL raises the gain from 1.84 to 6.66 points and reaches the strongest results consistently on all seven benchmarks, with the fewest steering actions and executor turns.
These results show that reinforcement learning effectively unlocks the steering intelligence, enabling the steering agent to mitigate goal drift.

Regarding the learning algorithm, we additionally compare online reinforcement learning with an offline alternative, where filtered behavior cloning (BC)~\citep{chen2021decision,yuan2023scaling} imitates the sampled steering actions that improve the task outcome. Filtered BC falls behind our online algorithm, further validating the necessity of online reinforcement learning.

\subsection{Generalizability, Transferability, and Controllability (RQ2)}

\begin{table}[t]
\centering
\footnotesize
\setlength{\tabcolsep}{4.5pt}
\vspace{-1.5em}
\caption{\textbf{Results on out-of-domain benchmarks with an online search API.} We report mean (Avg@4) and best-of-n (Pass@4) accuracy across 4 random seeds. Best results are in bold.}
\label{tab:ood}
\begin{tabular}{l|cc|cc|cc}
\toprule
\multirow{2}{*}{Method} & \multicolumn{2}{c|}{Frames}
 & \multicolumn{2}{c|}{GAIA}
 & \multicolumn{2}{c}{xBench-DS} \\
\cmidrule(lr){2-3}\cmidrule(lr){4-5}\cmidrule(lr){6-7}
 & Avg@4 & Pass@4 & Avg@4 & Pass@4 & Avg@4 & Pass@4 \\
\midrule
No Intervention & 26.46 & 44.17 & 16.75 & 34.95 & \textbf{49.50} & 65.00 \\
\ours Steering w/ RL & \textbf{36.32} & \textbf{59.71} & \textbf{23.79} & \textbf{39.81} & \textbf{49.50} & \textbf{72.00} \\
\bottomrule
\end{tabular}
\end{table}

To study whether the learned steering intelligence generalizes to out-of-domain tasks, we further evaluate on tasks that require an online search API, which exposes more complex real-world information.
We test the same steering model without additional training on three more challenging benchmarks that require deeper multi-hop search, namely Frames~\citep{krishna2025frames}, GAIA~\citep{mialon2023gaia}, and xBench-DeepSearch~\citep{chen2025xbench} (Table~\ref{tab:ood}). Steering improves both Avg@4 and Pass@4 on Frames and GAIA, and raises Pass@4 on xBench-DeepSearch. These results show that the learned steering capability generalizes beyond the training distribution. The steering agent mainly learns to recognize and correct goal drift, while task-specific ability stays within the frozen executor, which mitigates overfitting to the training tasks.

\begin{table}[t]
\centering
\footnotesize
\setlength{\tabcolsep}{3pt}
\vspace{-1.5em}
\caption{\textbf{Transferability of the learned steering intelligence to unseen frozen executors.} $\Delta$ is the gain in Avg. over No Intervention with the same executor. Best results are in bold.}
\label{tab:transfer}
\resizebox{\linewidth}{!}{%
\begin{tabular}{l|ccc|ccc|ccc|ccc}
\toprule
\multirow{2}{*}{Method} & \multicolumn{3}{c|}{Qwen2.5-7B-Instruct}
 & \multicolumn{3}{c|}{Llama-3.1-8B-Instruct}
 & \multicolumn{3}{c|}{Mistral-7B-Instruct-v0.3}
 & \multicolumn{3}{c}{Phi-4 (14B)} \\
\cmidrule(lr){2-4}\cmidrule(lr){5-7}\cmidrule(lr){8-10}\cmidrule(lr){11-13}
 & Avg. & $\Delta$ & Turns
 & Avg. & $\Delta$ & Turns
 & Avg. & $\Delta$ & Turns
 & Avg. & $\Delta$ & Turns \\
\midrule
No Intervention & 42.70 & \na & 6.4 & 33.14 & \na & 11.8 & 27.58 & \na & 7.8 & 35.35 & \na & 6.1 \\
Zero-Shot Steering & 44.54 & +1.84 & 6.5 & 35.66 & +2.52 & 11.1 & 29.31 & +1.73 & 7.7 & 35.98 & +0.63 & 6.0 \\
\ours RL Steering & \textbf{49.36} & \textbf{+6.66} & \textbf{6.2} & \textbf{48.92} & \textbf{+15.78} & \textbf{9.4} & \textbf{39.59} & \textbf{+12.01} & \textbf{6.9} & \textbf{41.27} & \textbf{+5.92} & \textbf{5.8} \\
\bottomrule
\end{tabular}%
}
\end{table}

To test whether the steering intelligence learned with one execution model can transfer across different model families, we use the same steering model to supervise three unseen executors, Llama-3.1-8B-Instruct~\citep{grattafiori2024llama}, Mistral-7B-Instruct-v0.3~\citep{jiang2023mistral} and Phi-4~\citep{abdin2024phi4}.
As shown in Table~\ref{tab:transfer}, zero-shot steering behaves consistently across the four model families and brings small gains.
The steering agent trained with the Qwen executor shows strong transferability, improving the three unseen executors by 15.78, 12.01, and 5.92 points while effectively reducing the executor turns.
These results show that the learned steering intelligence transfers to executors unseen during training, allowing the executor to be freely switched without retraining.

\begin{figure}[t]
\centering
\includegraphics[width=0.95\linewidth]{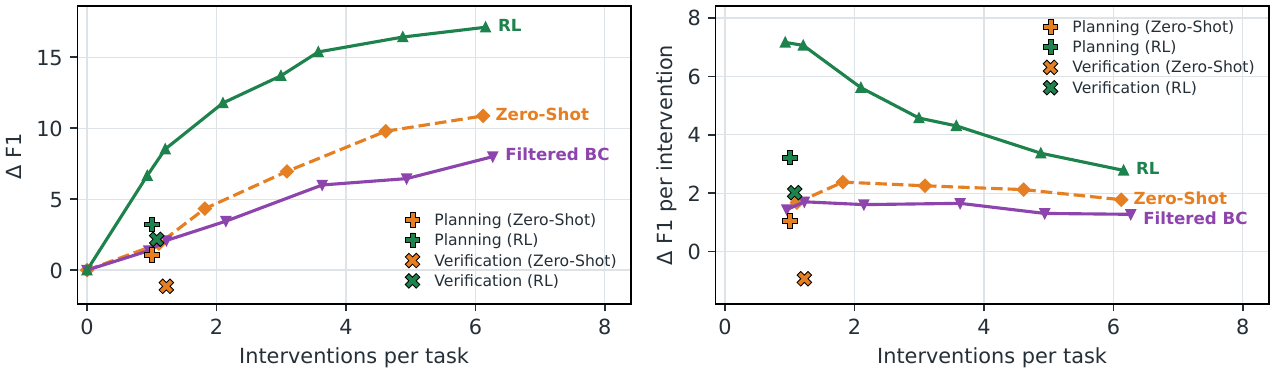}
\vspace{-1em}
\caption{\textbf{F1 improvement as the intervention frequency varies.} By adjusting the fast path threshold without retraining, RL steering maintains the largest total gain (left) and the largest gain per intervention at low frequency (right), prioritizing the states that influence the final outcome most.}
\label{fig:dose}
\vspace{-1em}
\end{figure}

We further examine the controllability of the intervention frequency by varying the threshold of Equation~\ref{eq:steering_gate} at deployment without retraining. We visualize the total gain and the gain per intervention in Figure~\ref{fig:dose}.
In the left panel, RL steering consistently improves task performance across all intervention budgets, surpassing zero-shot steering, Filtered BC, and the planning and verification baselines. At the highest intervention frequency, RL steering raises the total gain to 17.11 points.
These results show that the fast path design turns intervention frequency into a deployment-time control, allowing the same model to trade steering cost for task performance.
In the right panel, RL steering achieves its largest gain per intervention at low intervention frequency, while zero-shot steering and filtered BC remain low across frequencies, lacking the ability to judge where intervention matters most.
This demonstrates that the fast path learns to prioritize steering at the states that influence the final outcome most, yielding the largest gain under a fixed intervention budget.

\subsection{Steering Multi-Agent Systems (RQ3)}
\label{sec:mas}

\begin{table}[t]
\centering
\footnotesize
\setlength{\tabcolsep}{3pt}
\caption{\textbf{Results on the multi-agent setting.} $\Delta$ is the gain in Avg. over No Intervention, and Cost reports steering actions (\#Act.) and agent turns (Turns) per task. Best results are in bold.}
\label{tab:mas}
\resizebox{\linewidth}{!}{%
\begin{tabular}{l|cccc|ccc|cc|cc}
\toprule
\multirow{2}{*}{Method} & \multicolumn{4}{c|}{Multi-Hop QA}
 & \multicolumn{3}{c|}{Single-Hop QA}
 & \multicolumn{2}{c|}{Overall}
 & \multicolumn{2}{c}{Cost} \\
\cmidrule(lr){2-5}\cmidrule(lr){6-8}\cmidrule(lr){9-10}\cmidrule(lr){11-12}
 & 2Wiki & Hotpot & Bamboogle & MuSiQue & NQ & Trivia & Pop
 & Avg. & $\Delta$ & \#Act. & Turns \\
\midrule
No Intervention & 38.8 & 50.0 & 35.0 & 19.2 & \textbf{48.7} & 59.7 & \textbf{46.0} & 42.49 & \na & \na & \textbf{9.2} \\
\midrule
\multicolumn{12}{c}{\textit{Target frequency at 25\% turns}} \\
\midrule
\sub{Zero-Shot} & 42.8 & 47.8 & 36.7 & 20.8 & 40.1 & \textbf{61.3} & 40.6 & 41.44 & $-1.04$ & 3.37 & 13.5 \\
\ours \sub{w/ RL} & 50.6 & \textbf{53.2} & 47.8 & \textbf{24.0} & 45.1 & 60.3 & 43.7 & 46.39 & +3.90 & \textbf{2.67} & 10.4 \\
\midrule
\multicolumn{12}{c}{\textit{Steer at every handoff}} \\
\midrule
\sub{Zero-Shot} & 42.8 & 39.6 & 42.2 & 15.4 & 28.6 & 49.0 & 29.8 & 35.34 & $-7.14$ & 10.19 & 24.1 \\
\ours \sub{w/ RL} & \textbf{56.9} & 53.1 & \textbf{51.3} & \textbf{24.0} & 45.2 & 61.1 & 43.2 & \textbf{47.83} & \textbf{+5.34} & 6.00 & 11.9 \\
\bottomrule
\end{tabular}%
}
\end{table}

We further conduct experiments on the multi-agent setting. Following TacoMAS~\citep{xu2026tacomas}, we build a five-role pipeline for search tasks, where agents with specialized roles collaborate in a fixed order, details of which are provided in Appendix~\ref{app:mas}. The observation space of the steering agent is the task goal and inter-agent handoffs. The action space includes appending a guidance note or rejecting the message for retry. The steering agent, the training procedure, and the benchmarks stay the same as in the single-agent experiments. Table~\ref{tab:mas} reports the F1 metrics, with cost measured by steering actions (\#Act.) and agent turns (Turns), the total number of actions executed by agents in the MAS.

Results show that supervising an MAS with lower granularity is more challenging for the steering agent. With the threshold of Equation~\ref{eq:steering_gate} tuned so that steering actions cover 25\% of agent turns, zero-shot steering lowers the average F1 by 1.04 points. Forcing steering at every handoff increases the drop to 7.14 points, with agent turns inflated from 9.2 to 24.1 per task.
In comparison, our RL algorithm effectively improves the steering intelligence, achieving gains of 3.90 and 5.34 points under the two frequency settings, respectively.
These results show that the same steering interface and learning algorithm are applicable to both fine-grained supervision of a single agent and coarse-grained supervision of a multi-agent system.

\section{Conclusion}
This paper targets goal drift in LLM-based agent systems and builds a steering agent that decides when to intervene with a low-cost fast path and how with a deliberative slow path. Joint reinforcement learning over the two paths unlocks steering intelligence that improves both single-agent and multi-agent systems, generalizes to out-of-domain tasks, transfers to unseen executors, and supports controllable intervention frequency. As LLM-based agents and multi-agent systems tackle increasingly complex and long-horizon tasks, the steering agent provides an effective on-the-fly supervision tool to improve their performance and efficiency.

\bibliography{iclr2027_conference}
\bibliographystyle{iclr2027_conference}

\ifshowappendix
\newpage
\appendix
\section{Implementation Details}
\label{app:implementation}

\subsection{Training Details}
\label{app:training_details}

Single agent and multi-agent experiments share the same training setup. During training, the steering agent is initialized from Qwen3-8B~\citep{yang2025qwen3}. The backbone of the execution agent system is Qwen2.5-7B-Instruct~\citep{qwen2024qwen25}, which is served separately and excluded from parameter updates. We train with AdamW~\citep{loshchilov2019adamw} at a learning rate of $2\times10^{-5}$ and a batch size of 64 tasks for 300 steps.
The task reward $R$, first defined in Section~\ref{sec:protocol}, is instantiated as answer token F1 multiplied by a trajectory-format indicator. The F1 term measures the word-level overlap between the predicted and the reference answers, and the indicator is 1 when every executor turn commits exactly one well-formed action tag and 0 otherwise.
The remaining reinforcement learning hyperparameters follow the defaults of the ASearcher framework.

Each task performs $m=2$ selection rounds over the state bank, and each round selects up to $k=3$ states. Each selection takes a uniformly random state with probability $\varepsilon_{\mathrm{sel}}=0.25$ and otherwise the state with the highest steering priority.
Every selected state receives its full group of $G=4$ steering actions. Each steered and unsteered state is evaluated by $N=2$ executor continuations, whose means give $\bar R_{s,j}$ and the baseline $b_s$ in Equation~\ref{eq:outcome_stats}. All new states reached by the continuations are added to the state bank.

We train on ASearcher-Base-35k~\citep{gao2025asearcher}, a challenging QA dataset constructed from filtered open-source multi-hop questions and agent-synthesized QA pairs. The training environment is the same local retrieval server as the in-domain evaluation, using the 2018 Wikipedia dump~\citep{karpukhin2020dense} as the information source and E5~\citep{wang2022text} as the retriever.

\subsection{Evaluation Details}
\label{app:eval_config}

We follow ASearcher~\citep{gao2025asearcher} to report token F1 on the in-domain benchmarks, and Avg. is the mean over the seven benchmarks. Out-of-domain evaluation runs every question 4 times and reports Avg@4, the mean accuracy over the four runs, and Pass@4, the fraction of questions solved in at least one run. We follow ASearcher to judge answer correctness on these benchmarks with an LLM (Qwen3.8-max~\citep{yang2025qwen3}) as judge, and the judge prompt template is listed in Appendix~\ref{app:prompts}.

The in-domain evaluation data of ASearcher~\citep{gao2025asearcher} includes seven retrieval-augmented QA benchmarks, totaling 6,125 questions with 1,000 sampled questions from each benchmark and the full 125 questions of Bamboogle.
2WikiMultiHopQA~\citep{ho2020constructing} builds multi-hop questions from Wikipedia and Wikidata with annotated reasoning paths. HotpotQA~\citep{yang2018hotpotqa} contains two-hop questions that bridge or compare Wikipedia entities. Bamboogle~\citep{press2023measuring} collects hand-written compositional questions that a standard search engine answers incorrectly. MuSiQue~\citep{trivedi2022musique} composes questions of two to four hops from single-hop ones with stricter connectivity requirements. Natural Questions~\citep{kwiatkowski2019natural} contains real search-engine queries answered by Wikipedia. TriviaQA~\citep{joshi2017triviaqa} contains trivia questions authored by quiz enthusiasts. PopQA~\citep{mallen2023popqa} probes long-tail entity knowledge with templated questions built from Wikidata triples.

The out-of-domain evaluation uses three benchmarks that require deeper search over the open web. Frames~\citep{krishna2025frames} contains 824 questions that combine multiple constraints and typically require evidence from several Wikipedia articles. GAIA~\citep{mialon2023gaia} is a general assistant benchmark, from which we evaluate the 103 text-only questions of its validation split. xBench-DeepSearch~\citep{chen2025xbench} contains 100 questions targeting the deep-search ability of agents. These evaluations use an online search engine.

\subsection{Baseline Design}
\label{app:baselines}

The planning, verification, and filtered behavior cloning baselines use the same supervisor backbone, and the trained variants reuse the reinforcement learning algorithm, data, and hyperparameters for a fair comparison. The comparison therefore isolates the supervision placement and the learning algorithm.

\paragraph{Planning.}
The planning agent is triggered once before the executor's first action. It reads the same task requirement as the steering agent and writes a plan that the executor keeps in view for the whole task.

\paragraph{Verification.}
The verification agent is triggered after the executor commits an answer and decides whether the submission stands. It either passes the answer or rejects it with feedback. The feedback is read by the executor agent as a tool-return, and the executor revises and resubmits. After three rejections the latest answer is delivered without further review.

\paragraph{Filtered behavior cloning.}
The filtered BC baseline~\citep{wang2020critic,chen2021decision} collects the steering actions sampled by the same rollout pipeline as our online RL algorithm and keeps those whose mean reward $\bar R_{s,j}$ exceeds the unsteered baseline $b_s$ in Equation~\ref{eq:outcome_stats}.
Then it treats these positive samples as the training data for supervised fine-tuning (SFT)~\citep{ouyang2022instructgpt}.
The retained pairs of observation and steering action fine-tune the steering agent with a token-level cross-entropy objective~\citep{zelikman2022star,yuan2023scaling}. Compared with our online algorithm, filtered BC is an offline algorithm, so it receives no signal from the discarded negative samples and its training data does not track the improving policy.

\subsection{Multi-Agent Instantiation}
\label{app:mas}

The multi-agent system adapts the five-role template of TacoMAS~\citep{xu2026tacomas} to search tasks. A planner decomposes the question into subgoals and dispatches them to two searchers, which retrieve evidence in parallel. A verifier reviews the searchers' deliverables against the question, and a summarizer composes the final answer from the delivered evidence and the verifier's verdict.

The steering agent reviews a pending handoff using the task, the multi-agent structure, previously delivered handoffs, current handoff, its sender and recipients, and any rejection feedback already issued at that handoff.
The action space contains a pass that delivers the handoff unchanged, a note that attaches guidance to the delivered submission for its recipients, and a retry that returns rejection feedback to the sender for revision and resubmission. The full prompt templates are listed in Appendix~\ref{app:prompts}.

\section{Additional Experimental Results}
\label{app:experiments}

\subsection{Hyperparameters}
To study the impact of hyperparameters, we retrain the steering agent varying the number of selection rounds $m$, the states selected per round $k$, and the continuations per branch $N$, keeping the group size $G=4$ fixed. We evaluate the retrained agents under the in-domain protocol of Table~\ref{tab:id} at three target steering frequencies. As shown in Figure~\ref{fig:hyper-parameters}, all configurations deliver comparable gains at the 15\% frequency, staying within a band of one F1 point. When the frequency increases to 50\% and 100\%, other configurations demonstrate lower performance.
\begin{figure}[t]
    \centering
    \includegraphics[width=0.85\linewidth]{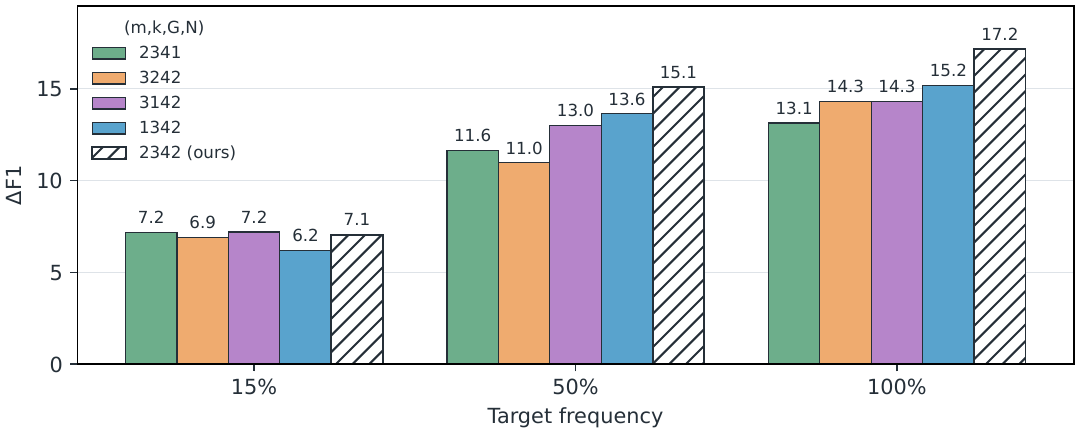}
    \caption{\textbf{In-domain gains under varied rollout hyperparameters.} Each configuration is labelled by $(m,k,G,N)$, the number of selection rounds, states per round, steering actions per state, and continuations per branch. $\Delta$F1 is the paired gain in mean F1 over the seven benchmarks against the no-intervention baseline.}
    \label{fig:hyper-parameters}
\end{figure}

\subsection{Cost of the Fast and Slow Paths}
\label{app:fastslow_latency}

\begin{figure}[t]
    \centering
    \includegraphics[width=0.42\linewidth]{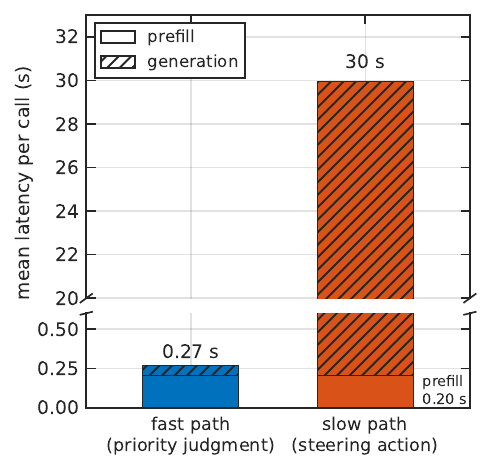}
    \caption{\textbf{Average latency per call of the fast and slow paths.} The two paths share the same prefill cost, and the cost of the slow path comes from generating the reasoning and steering action.}
    \label{fig:fastslow_latency}
\end{figure}

To examine the actual cost of the low-cost fast path and the high-cost slow path, we measure their latency on the trained steering agent served by SGLang~\citep{zheng2024sglang}. We collect the latency data during the in-domain evaluation of Table~\ref{tab:id} and report the average latency per call at 12 concurrent requests per GPU.
As shown in Figure~\ref{fig:fastslow_latency}, the two paths read the same context and therefore share the same prefill cost of about 0.2 s. The fast path then decodes the decision token and finishes within 0.3 s, while the slow path generates about 1.8k reasoning and action tokens on average and takes 30 s. These results show that the fast-slow design effectively separates the cost of judgment from the cost of intervention, allowing the fast path to run at every observation update while the slow path conducts reasoning only when triggered.

\subsection{LLM-as-a-Judge Results on In-Domain Benchmarks}
\label{app:id_lasj}

\begin{table}[t]
\centering
\footnotesize
\setlength{\tabcolsep}{3pt}
\caption{\textbf{LLM-as-a-Judge results on in-domain benchmarks.} The methods and predictions are the same as in Table~\ref{tab:id}, and $\Delta$ is the gain in Avg. over No Intervention. Best results are in bold.}
\label{tab:id_lasj}
\begin{tabular}{l|cccc|ccc|cc}
\toprule
\multirow{2}{*}{Method} & \multicolumn{4}{c|}{Multi-Hop QA}
 & \multicolumn{3}{c|}{Single-Hop QA}
 & \multicolumn{2}{c}{Overall} \\
\cmidrule(lr){2-5}\cmidrule(lr){6-8}\cmidrule(lr){9-10}
 & 2Wiki & Hotpot & Bamboogle & MuSiQue & NQ & Trivia & Pop
 & Avg. & $\Delta$ \\
\midrule
No Intervention & 44.6 & 55.7 & 38.4 & 17.0 & 53.7 & 72.8 & 47.4 & 47.09 & \na \\
\midrule
Planning & 44.6 & 53.8 & 48.0 & 20.4 & 55.2 & 74.7 & 49.3 & 49.43 & +2.34 \\
\sub{w/ RL} & 48.4 & 53.5 & 45.6 & 18.8 & 52.0 & 72.4 & 47.4 & 48.30 & +1.21 \\
\midrule
Verification & 42.8 & 52.2 & 45.6 & 16.2 & 50.8 & 74.2 & 43.0 & 46.40 & $-0.69$ \\
\sub{w/ RL} & 43.1 & 55.1 & 42.4 & 19.7 & 56.3 & 71.7 & 47.7 & 48.00 & +0.91 \\
\midrule
Steering & 49.3 & 55.9 & 39.2 & 22.7 & 54.8 & 72.2 & 46.7 & 48.69 & +1.60 \\
\sub{w/ Filtered BC} & 50.3 & 55.5 & 40.0 & 21.4 & 55.5 & 73.8 & 48.5 & 49.29 & +2.20 \\
\ours \sub{w/ RL} & \textbf{56.2} & \textbf{60.5} & \textbf{49.6} & \textbf{24.4} & \textbf{58.4} & \textbf{76.0} & \textbf{51.9} & \textbf{53.86} & \textbf{+6.77} \\
\bottomrule
\end{tabular}
\end{table}

To examine whether the in-domain results depend on the token F1 metric, we follow ASearcher~\citep{gao2025asearcher} to supplement LLM-as-a-Judge (LasJ) accuracy for all methods in Table~\ref{tab:id}. The judge model and prompt template are the same as in the out-of-domain evaluation (Appendix~\ref{app:eval_config}). As shown in Table~\ref{tab:id_lasj}, LasJ accuracy exhibits the same trend as Table~\ref{tab:id}. The RL improved steering agent achieves the best results consistently on all seven benchmarks, and RL raises the gain of steering from 1.60 to 6.77 points. LasJ accuracy is higher than token F1 overall, as the judge accepts answers that match the reference in meaning but differ in surface form.

\section{Prompt Templates}
\label{app:prompts}

In this section, we list the full prompt templates used in the experiments. Curly-braced slots are filled at run time from the current observation. In the single-agent setting, the slots carry the question, the executor's rendered action history, the action count with the remaining turn budget, and a line naming the executor's currently legal actions. A committed note is injected into the executor's prompt under the header ``A steering supervisor has reviewed your progress and left a note for your next action (follow the note):''. In the multi-agent setting, the slots carry the task, the rendered pipeline, the handoffs delivered so far, the submission under review with its sender and recipients, and any rejection feedback already issued at this handoff. The planning and verification prompts share the single-agent context, with the role statement adapted to their placement. The last box lists the LLM-as-Judge prompt of the out-of-domain evaluation.

\begin{promptbox}{Single-agent fast-path prompt}
You are a steering supervisor watching over an autonomous search agent. The agent solves a question with a search engine; you see the question and its full action history, exactly as the agent sees them. You may send the agent a short steering note. A note is shown to the agent for its next action only, then disappears.

Question:
```txt
{question}
```

Agent's action history:
```txt
{history}
```

Agent status: it has taken {num_actions} actions; at most {remaining_turns} turns remain before it must answer.
{action_space}

Decide whether this is a critical state. Criticality is the degree to which the agent's next action influences the final outcome: at a critical state, different choices of the next action lead to meaningfully different final results; at a non-critical state, the next action has little influence on how things end.

Reply with exactly one of:
<judge>yes</judge> -- high criticality: the next action strongly influences the final outcome
<judge>no</judge> -- low criticality: the next action has little influence on the final outcome
\end{promptbox}

\begin{promptbox}{Single-agent slow-path prompt}
You are a steering supervisor watching over an autonomous search agent. The agent solves a question with a search engine; you see the question and its full action history, exactly as the agent sees them. You may send the agent a short steering note. A note is shown to the agent for its next action only, then disappears.

Question:
```txt
{question}
```

Agent's action history:
```txt
{history}
```

Agent status: it has taken {num_actions} actions; at most {remaining_turns} turns remain before it must answer.
{action_space}

You have decided to intervene at this point. Think through what is going wrong or what opportunity the agent is missing, then write a steering note for the agent.

The note will be shown to the agent for its NEXT action only, then disappear. Make it immediately actionable for that one decision: name the exact search query to run, the specific evidence to re-read, or the concrete correction to apply now.

After your reasoning, reply with the note inside tags:
<note>your steering note (1-3 concise sentences, actionable this turn)</note>
\end{promptbox}

\begin{promptbox}{Multi-agent fast-path prompt}
You are a steering supervisor watching over a multi-agent system that is solving a task. The system is a fixed pipeline of role agents. One agent has just finished, and its output is about to be handed to the downstream agent(s).

Task:
```txt
{task}
```

Pipeline:
```txt
{pipeline}
```

{history_block}

Handoff under review:
- from: {from_desc}
- to: {to_desc}
- position: {position}
- submission:
```txt
{submission}
```

Retries used at this handoff: {retries_used} of {max_retries}.{retry_feedback_block}

Decide whether this handoff is a critical state. Criticality is the degree to which intervening on this handoff influences the final outcome: at a critical state, rejecting or annotating this submission leads to a meaningfully different final result; at a non-critical state, passing it through has little influence on how things end.

Reply with exactly one of:
<judge>yes</judge> -- high criticality: intervening on this handoff strongly influences the final outcome
<judge>no</judge> -- low criticality: this handoff can pass as-is with little influence on the final outcome
\end{promptbox}

\begin{promptbox}{Multi-agent slow-path prompt}
You are a steering supervisor watching over a multi-agent system that is solving a task. The system is a fixed pipeline of role agents. One agent has just finished, and its output is about to be handed to the downstream agent(s).

Task:
```txt
{task}
```

Pipeline:
```txt
{pipeline}
```

{history_block}

Handoff under review:
- from: {from_desc}
- to: {to_desc}
- position: {position}
- submission:
```txt
{submission}
```

Retries used at this handoff: {retries_used} of {max_retries}.{retry_feedback_block}

You have decided to intervene at this handoff. Think through what is wrong with the submission or what the receiving agent needs to know, then commit exactly one action:

<retry>your rejection feedback -- the submission is rejected and the rejection is returned to the submitting agent as the result of its submission; it will redo the work with your feedback in view</retry>
<note>your note -- the submission passes through unchanged, and this note is attached to it for the receiving agent(s)</note>
\end{promptbox}

\begin{promptbox}{Planning prompt}
You are a steering supervisor watching over an autonomous search agent. The agent solves a question with a search engine; you see exactly what the agent sees: the question and its action history so far. The agent is about to decide its next action.

Question:
```txt
{question}
```

Agent's action history:
```txt
{history}
```
{prior_notes_block}
Agent status: it has taken {num_actions} actions; at most {remaining_turns} turns remain before it must answer.
{action_space}

The agent has not acted yet. Think through how it should approach this question, then write a steering note that the agent will keep in view for the whole task.

After your reasoning, reply with the note inside tags:
<note>your steering note (1-3 concise sentences: point out the problem and what to do next)</note>
\end{promptbox}

\begin{promptbox}{Verification prompt}
You are a steering supervisor watching over an autonomous search agent. The agent solves a question with a search engine; you see the question and its full action history, exactly as the agent sees them. The agent has just submitted an answer; you decide whether it stands.

Question:
```txt
{question}
```

Agent's action history:
```txt
{history}
```

Agent status: it has taken {num_actions} actions; at most {remaining_turns} turns remain before it must answer.
{action_space}

The agent has submitted the answer shown as the last turn of its history. Decide whether to accept it. Think it through, then commit exactly one action:

<pass></pass> -- the answer is delivered unchanged, no intervention
<retry>your rejection feedback -- the submission is rejected and the rejection is returned to the agent as the result of its submission; it will redo the work with your feedback in view</retry>
\end{promptbox}

\begin{promptbox}{LLM-as-Judge prompt}
You are an evaluation assistant. Please determine if the predicted answer is equivalent to the labeled answer.

Question: {question}

Labeled Answer: {gt_answer}

Predicted Answer: {pred_answer}

Did the model give an answer **equivalent** to the labeled answer? Please respond with "Correct" if they are equivalent, or "Incorrect" if they are not equivalent.

The output should in the following json format:
```json
{
    "rationale": your rationale for the judgement, as a text,
    "judgement": your judgement result, can only be "Correct" or "Incorrect",
}
```
\end{promptbox}

\fi

\end{document}